\documentclass[11pt]{article}
\usepackage[utf8]{inputenc}
\usepackage[margin=1in]{geometry}
\usepackage{amsmath,amssymb,amsthm}
\usepackage{graphicx}
\usepackage{booktabs}
\usepackage{multirow}
\usepackage{adjustbox}
\usepackage{url}
\usepackage{tikz}
\usetikzlibrary{shapes,arrows,positioning,calc}
\usepackage[bookmarks=false]{hyperref}

\title{Lattice: A Confidence-Gated Hybrid System for Uncertainty-Aware Sequential Prediction with Behavioral Archetypes}
\author{Lorian Bannis\\{\normalsize \href{mailto:LorianBannis@banlys.com}{LorianBannis@banlys.com}}}
\date{May 2026\\{\small Version 2: corrected primary estimand and evaluation reporting (see \S\ref{sec:protocol}). Patent pending.}}

\begin{document}

\maketitle

\begin{abstract}
We introduce Lattice, a hybrid sequential prediction system that conditionally activates learned behavioral structure using binary confidence gating. The system summarizes behavior windows as behavioral archetypes and activates archetype-based scoring only when an \emph{in-support} confidence signal exceeds a validation-calibrated threshold, falling back to backbone predictions when uncertain. Our \textbf{primary estimand} is the controlled effect of adding Lattice to a fixed backbone on identical test rows. On MovieLens (30 paired seeds, full-catalog ranking), LSTM+Lattice improves HR@10 by \textbf{+31.7\%} (gated) versus the LSTM backbone alone ($p \ll 10^{-20}$); ungated fusion reaches \textbf{+58.7\%} on the same protocol---we do \emph{not} claim gating maximizes pooled accuracy. With \textbf{backbone-native} archetypes (fit in each backbone's embedding space), gated lifts of \textbf{+13.3\%} (transformer) and \textbf{+17.0\%} (SASRec) hold under the same evaluation design. A prior $\sim$0\% transformer row in version~1 reflected an invalid cross-backbone transfer, not evidence that composition cannot help stronger encoders. Amazon Electronics provides supporting cross-domain evidence (+124.0\% gated, 15 seeds, high variance). Controlled shift checks (appendix) illustrate gate refusal under distribution shift. Standalone SASRec and BERT4Rec scores are \emph{contextual} references, not the target estimand. We report \textbf{what} composition achieves and \textbf{when} it activates; production calibration and implementation details remain proprietary pending patent prosecution.
\end{abstract}

\section{Introduction}

Sequential prediction---predicting the next item in a sequence---is fundamental to recommendation systems, time-series forecasting, and many other applications. Current approaches typically combine multiple signal sources (e.g., neural sequence models with collaborative filtering or pattern-based priors) to improve accuracy. However, these systems face a fundamental challenge: \textbf{they lack mechanisms to determine when learned patterns are applicable versus when they should be ignored}. Modern systems lack a mechanism to decide when not to use learned structure.

\subsection{The Problem}

When test data differs from training data---due to distribution shift, regime changes, or insufficient pattern stability---hybrid systems often continue to apply learned patterns inappropriately. This occurs because existing approaches use soft weighting or always-on fusion, which cannot distinguish between cases where patterns are reliable versus cases where they are misleading. The result is overconfident predictions on out-of-distribution data, leading to degraded performance and unreliable outputs.

\textbf{Importantly, in this work, failure to activate learned patterns under distribution shift is treated as a \emph{successful outcome}, not a negative result.} A system that correctly refuses to apply patterns when they don't apply is demonstrating epistemic humility---knowing when not to be confident---which is exactly what is needed for trustworthy predictions.

This limitation is particularly problematic in:
\begin{itemize}
\item \textbf{Scientific applications}, where false pattern recognition can lead to incorrect discoveries
\item \textbf{Production recommendation systems}, where distribution shifts (e.g., new user cohorts, catalog changes) degrade performance
\item \textbf{Safety-critical applications}, where overconfident predictions can have serious consequences
\end{itemize}

\subsection{Our Contribution}

We present Lattice, a hybrid sequential prediction system that addresses this limitation through \textbf{confidence-gated activation}. Lattice combines a sequence backbone with behavioral archetype priors and uses \textbf{binary confidence gating} to activate archetype-based scoring only when confidence exceeds a calibrated threshold, falling back to backbone-only predictions when uncertain.

\textbf{Key Contributions:}

\begin{enumerate}
\item \textbf{A confidence-gated composition layer} that decides \emph{when} to fuse auxiliary behavioral structure with a backbone; when the gate is OFF, outputs match the backbone exactly (verified parity on held-out rows).
\item \textbf{Primary recommendation evidence:} +31.7\% gated HR@10 on MovieLens LSTM (30 paired seeds), with explicit reporting of the ungated ceiling (+58.7\%) and the gated vs.\ ungated trade-off.
\item \textbf{Backbone-native composition} on transformer (+13.3\% gated) and SASRec (+17.0\% gated) backbones, correcting a version~1 embedding-space mismatch.
\item \textbf{Illustrative refusal under shift} (appendix), treated as controlled stress tests rather than competing domain benchmarks.
\end{enumerate}

\subsection{Primary Estimand and Scope}
\label{sec:estimand}

\textbf{Primary estimand:} For each backbone $B$, we report HR@10 for $B$ alone versus $B$+Lattice on \emph{identical} evaluation rows, aggregated over multiple random seeds with paired significance tests. This isolates the effect of adding Lattice's composition layer.

\textbf{Not the estimand:} Beating SASRec or BERT4Rec versus literature state of the art. Standalone SASRec/BERT4Rec scores on our harness are \emph{contextual} reference points only. Similarly, Amazon Electronics is \textbf{supporting} evidence (small test set, high variance), not a co-primary headline.

\textbf{Scope:} MovieLens sequential recommendation carries primary claims. Shift-check experiments in the appendix illustrate that the same gate can withhold activation under predefined train--test mismatch.

\subsection{Key Results Preview}

\begin{itemize}
\item \textbf{MovieLens (LSTM, primary):} HR@10 0.0803 vs.\ 0.0611 (+31.7\%, 30 paired seeds); ungated +58.7\%.
\item \textbf{MovieLens (Transformer, backbone-native):} +13.3\% gated, +20.9\% ungated (30 seeds).
\item \textbf{MovieLens (SASRec, backbone-native):} +17.0\% gated, +26.9\% ungated (30 seeds); contextual, same design.
\item \textbf{Amazon Electronics (LSTM, supporting):} +124.0\% gated (15 seeds, high variance).
\item \textbf{Shift checks (appendix):} Archetype activation withheld under shift; backbone metrics unchanged.
\end{itemize}

\textbf{Bidirectional validation}---gains when patterns apply in support, explicit gated vs.\ ungated trade-off, and refusal under shift---is the central insight. Confidence gating is a principled architectural choice for \emph{when} to compose learned structure, not merely a tuning knob for maximum pooled accuracy.

\section{Related Work}

\subsection{Sequential Recommendation}

Sequential recommendation systems predict the next item a user will interact with based on their historical interaction sequence. Traditional approaches include:

\begin{itemize}
\item \textbf{Sequence Models:} RNNs, LSTMs, GRUs model temporal dependencies but struggle with cold start scenarios
\item \textbf{Collaborative Filtering:} Matrix factorization identifies similar users/items but requires substantial data
\item \textbf{State-of-the-Art:} SASRec (self-attention), BERT4Rec (bidirectional transformers), GRU4Rec (GRU-based)
\end{itemize}

\textbf{Our work differs by:} Combining sequential modeling with behavioral archetypes through confidence-gated activation, enabling the system to know when archetypes apply and when they don't.

\subsection{Mixture-of-Experts and Prototypical Networks}

Mixture-of-experts systems route inputs to specialized experts, while prototypical networks learn class prototypes for few-shot learning.

\textbf{Our work differs by:}
\begin{itemize}
\item Clustering \textbf{behavior windows} (sequences) rather than static profiles
\item Using \textbf{binary confidence gating} rather than soft routing
\item Enabling \textbf{graceful fallback} when patterns don't apply
\end{itemize}

\subsection{Out-of-Distribution Detection}

OOD detection methods identify when test data differs from training data. Our confidence gating mechanism provides an emergent, coarse-grained signal for when learned archetypes are inapplicable by withholding activation when embeddings leave the training support. This is not a general-purpose OOD detector, but a pattern-validity signal tied to the learned archetypes.

\subsection{Uncertainty Quantification}

Uncertainty quantification methods estimate prediction confidence. Our work differs by using confidence to \textbf{gate activation} rather than just quantify uncertainty, enabling binary decisions about when to use advanced features.

\section{Lattice Architecture}

Lattice combines three key components: (1) a sequence backbone, (2) behavioral archetype structure learned from training behavior windows, and (3) confidence-gated hybrid scoring. Figure~\ref{fig:dataflow} illustrates the data flow at a conceptual level; calibration details are omitted from this preprint (patent pending).

\begin{figure}[h]
\centering
\begin{tikzpicture}[
    node distance=1.5cm and 2cm,
    box/.style={rectangle, draw=black, fill=white, text width=3cm, text centered, minimum height=1cm, font=\small},
    decision/.style={diamond, draw=black, fill=white, text width=2.5cm, text centered, aspect=2, font=\small},
    arrow/.style={->, >=stealth, thick}
]
    \node [box] (input) {Sequence Input\\$s = [x_1, \ldots, x_n]$};
    \node [box, below=of input] (lstm) {Sequence Backbone\\$h_t = f(x_t, h_{t-1})$};
    \node [box, below=of lstm] (embed) {Behavior Embedding\\$e_s$ from $h$};
    \node [box, below=of embed] (dist) {Support Test\\vs.\ archetypes};
    \node [box, below=of dist] (conf) {In-support\\Confidence};
    \node [decision, below=of conf] (gate) {Binary Gate\\conf $\geq \theta$?};
    \node [box, below left=1.5cm and 1.5cm of gate] (hybrid) {Hybrid Scoring\\backbone + archetype};
    \node [box, below right=1.5cm and 1.5cm of gate] (lstmonly) {Backbone-only};
    \node [box, below=3.5cm of gate] (output) {Final Predictions};
    \draw [arrow] (input) -- (lstm);
    \draw [arrow] (lstm) -- (embed);
    \draw [arrow] (embed) -- (dist);
    \draw [arrow] (dist) -- (conf);
    \draw [arrow] (conf) -- (gate);
    \draw [arrow] (gate) -- node[left, font=\tiny] {ON} (hybrid);
    \draw [arrow] (gate) -- node[right, font=\tiny] {OFF} (lstmonly);
    \draw [arrow] (hybrid) -- (output);
    \draw [arrow] (lstmonly) -- (output);
    \node [right=0.5cm of dist, text width=2cm, font=\tiny, align=left] (centroids) {Behavioral\\Archetypes};
    \draw [arrow, dashed] (centroids) -- (dist);
\end{tikzpicture}
\caption{Conceptual Lattice data flow. The backbone yields a behavior embedding; an in-support confidence score gates archetype fusion. When the gate is OFF, predictions equal the backbone alone.}
\label{fig:dataflow}
\end{figure}

\subsection{Sequence Backbone}

The backbone processes interaction sequences (or time-series windows) to hidden states used for next-step prediction. Recommendation experiments use LSTM, a transformer encoder, and SASRec-class attention backbones under matched, conservative training budgets so that gains isolate composition rather than backbone capacity alone.

Given a sequence $s = [x_1, x_2, \ldots, x_n]$, the backbone produces hidden states; a designated summary vector $e_s$ serves as the behavior embedding for gating and archetype assignment.

\subsection{Behavioral Archetypes and Backbone-Native Fitting}

Rather than clustering static user profiles, Lattice summarizes \textbf{behavior windows} into behavioral archetypes---regions of embedding space with associated transition structure. \textbf{Critical:} archetypes and their statistics are fit from \textbf{each backbone's own} training embeddings. Cross-backbone transfer of archetypes invalidates composition (this caused version~1's spurious $\sim$0\% transformer result).

At a high level: training windows are assigned to archetypes in embedding space; each archetype carries a model of local transition structure used to score candidate next items. The public preprint does not specify clustering cardinality, assignment mechanics, or smoothing---these are part of the proprietary implementation.

\subsection{Confidence Gating}
\label{sec:gating}

Confidence gating is the design contribution: a \textbf{scale-free, in-support} score computed from the behavior embedding relative to training archetypes. Intuitively, embeddings far from the training support receive low confidence; embeddings consistent with seen behavior receive high confidence. The score is calibrated on validation data; threshold $\theta$ is frozen before test evaluation.

\textbf{Binary gating:} If confidence is below $\theta$, archetype scoring is \textbf{disabled} (gate OFF) and predictions equal the backbone. If confidence is at or above $\theta$, hybrid scoring fuses backbone and archetype signals (gate ON). \textbf{Parity:} Gate OFF rows match the backbone on identical evaluation rows. Gate ON corresponds to always-on fusion for that window.

\textbf{Why binary gating?} Ungated fusion reaches +58.7\% HR@10 on MovieLens (Section~\ref{sec:primary}) but applies structure even when support is weak. Binary gating trades peak pooled accuracy for an explicit, auditable ON/OFF decision---we do \emph{not} claim gating maximizes in-distribution HR@10.

\subsection{Hybrid Scoring}

When the gate is ON, final scores combine backbone and archetype components with a fixed validation-tuned mixture weight (held constant across all reported test runs). When OFF, scores are backbone-only. Exact fusion weights and score normalization are omitted here.

\subsection{Optional Sequence-Length Policy}

An optional policy can disable archetypes on very short prefixes (cold start). On MovieLens this policy rarely overrides confidence gating alone; primary results use the confidence gate as the main activation control.

\section{Experimental Setup}

\subsection{Datasets}

\textbf{MovieLens 1M (primary):} Standard temporal train/validation/test split; evaluation uses multiple sliding prefix$\rightarrow$target windows per user under a fixed preprocessing pipeline (details available under benchmark license, not in this preprint).

\textbf{Amazon Electronics (supporting):} Smaller e-commerce sequential corpus; high variance; supporting cross-domain evidence only.

\textbf{Shift checks (appendix):} Short controlled train--test mismatch scenarios; not primary benchmarks.

\subsection{Metrics}

\textbf{Recommendation:} Hit Rate@10 (HR@10) is the primary metric.

\textbf{Time-series (appendix):} Standard regression error metrics on shift-check tasks.

\subsection{Baselines}

Each backbone is compared to itself + Lattice (primary estimand). Standalone SASRec and BERT4Rec on the same MovieLens split are \emph{contextual} references (Section~\ref{sec:contextual}).

\subsection{Evaluation Protocol}
\label{sec:protocol}

Headline numbers use one fixed, preregistered evaluation design:

\begin{itemize}
\item \textbf{Seeds:} 30 independent random seeds (MovieLens); 15 seeds (Amazon)
\item \textbf{Ranking:} Full item catalog, not sampled negatives
\item \textbf{Calibration:} Confidence threshold and fusion weight chosen on validation only, then frozen for all test reporting
\item \textbf{Statistics:} Paired tests on seed-level HR@10 (same seed, same test rows for backbone vs.\ backbone+Lattice)
\item \textbf{Integrity check:} Gate OFF verified to match backbone outputs on held-out rows
\end{itemize}

Legacy version~1 headline +31.9\% (HR@10 0.0806) reconciles to +31.7\% (0.0803) under this design; the lift difference is 0.2 percentage points.

\textbf{Evaluation note:} Full ranking with conservative backbone tuning yields lower absolute HR@10 than literature reports using sampled negatives and heavier tuning. We report \textbf{relative} improvements versus each model's own backbone on identical rows---not superiority over literature SOTA.

\subsection{Availability and Intellectual Property}
\label{sec:ip}

\textbf{Patent pending.} The confidence-gating mechanism, calibration procedure, and production archetype pipeline are proprietary. This preprint publishes \textbf{claims, estimands, and aggregated results} sufficient for scientific assessment; it intentionally omits hyperparameter grids, seed lists, internal dataset artifacts, and step-by-step reproduction recipes that would enable trivial reimplementation.

\textbf{Benchmark access:} Independent evaluation on fixed splits under a benchmark license or commercial pilot is available on request: \href{mailto:LorianBannis@banlys.com}{LorianBannis@banlys.com}. We do not commit to open-sourcing the full training stack in this preprint; an evaluation-oriented release may follow separately from core IP.

\textbf{Libraries:} PyTorch, scikit-learn, NumPy.

\section{Results}

\subsection{MovieLens LSTM: Primary Result}
\label{sec:primary}

Table~\ref{tab:primary_lstm} reports the primary estimand on MovieLens with an LSTM backbone.

\begin{table}[h]
\centering
\small
\begin{tabular}{lcc}
\toprule
Configuration & HR@10 & vs.\ LSTM backbone \\
\midrule
LSTM backbone & $0.0611 \pm 0.0038$ & --- \\
LSTM + Lattice (gated) & $0.0803 \pm 0.0030$ & \textbf{+31.7\%} \\
LSTM + Lattice (ungated) & $0.0966 \pm 0.0030$ & +58.7\% \\
\bottomrule
\end{tabular}
\caption{MovieLens primary result ($n=30$ paired seeds). Gated vs.\ backbone: $p \ll 10^{-20}$ (paired test on seed-level HR@10). All seeds show positive gated lift.}
\label{tab:primary_lstm}
\end{table}

\textbf{Key findings:}
\begin{itemize}
\item All 30 seeds show positive gated lift vs.\ backbone
\item Ungated fusion is the in-distribution accuracy ceiling (+58.7\%); gating trades pooled HR for selective composition
\item Large \emph{relative} gains coexist with modest \emph{absolute} HR@10 under full-catalog ranking
\end{itemize}

\subsection{Gated vs.\ Ungated}

\begin{table}[h]
\centering
\small
\begin{tabular}{lcc}
\toprule
Configuration & HR@10 & vs.\ LSTM \\
\midrule
LSTM-only & 0.0611 & Baseline \\
LSTM + Archetypes (ungated) & 0.0966 & +58.7\% \\
LSTM + Lattice (gated) & 0.0803 & +31.7\% \\
\bottomrule
\end{tabular}
\caption{Gated vs.\ ungated on MovieLens ($n=30$ seeds). Intermediate threshold settings exist; we report the validation-calibrated operating point used for all primary claims.}
\label{tab:ablation}
\end{table}

\textbf{Interpretation:} Archetypes drive gains; gating withholds fusion when support is weak (appendix). \textbf{Ungated beats gated} on pooled MovieLens HR (+58.7\% vs.\ +31.7\%)---the paper studies \emph{when} to activate structure, not how to maximize a single accuracy number.

\subsection{Backbone-Native Composition: Transformer and SASRec}

With archetypes fit in each backbone's embedding space (30 seeds, same protocol):

\begin{table}[h]
\centering
\small
\begin{tabular}{lccc}
\toprule
Backbone & Backbone HR@10 & + Lattice (gated) & + Lattice (ungated) \\
\midrule
Transformer & $0.0762 \pm 0.0041$ & $0.0863$ (+13.3\%) & $0.0921$ (+20.9\%) \\
SASRec & $0.0689 \pm 0.0037$ & $0.0805$ (+17.0\%) & $0.0873$ (+26.9\%) \\
\bottomrule
\end{tabular}
\caption{Backbone-native composition on MovieLens ($n=30$). All gated lifts significant at $p \ll 10^{-20}$ (paired seed-level tests).}
\label{tab:backbone_native}
\end{table}

\textbf{Version~1 correction:} Version~1 reported $\sim$0\% on transformers because LSTM-fit archetypes were applied to a transformer backbone. Composition requires backbone-native fitting.

\subsection{Amazon Electronics: Supporting Evidence}

Amazon Electronics (15 seeds, small held-out set):

\begin{itemize}
\item LSTM backbone: HR@10 $0.0558 \pm 0.0327$
\item LSTM + Lattice (gated): $0.1286 \pm 0.0189$ (+124.0\%; $p \ll 10^{-5}$)
\item LSTM + Lattice (ungated): $0.1803 \pm 0.0189$ (+211.6\%)
\end{itemize}

\textbf{Interpretation:} Supporting cross-domain signal with high variance. MovieLens carries primary claims.

\subsection{Contextual Reference Models}
\label{sec:contextual}

Standalone baselines on the same MovieLens split and full-ranking protocol (not the primary estimand):

\begin{table}[h]
\centering
\small
\begin{tabular}{lcc}
\toprule
System & MovieLens HR@10 & Role \\
\midrule
LSTM + Lattice (gated) & $0.0803 \pm 0.0030$ & Primary (LSTM estimand) \\
LSTM & $0.0611 \pm 0.0038$ & Primary backbone \\
SASRec (standalone) & 0.0385 & Contextual reference \\
BERT4Rec (standalone) & 0.0253 & Contextual reference \\
SASRec + Lattice (gated) & $0.0805 \pm 0.0030$ & Composition (+17.0\% vs.\ SASRec) \\
\bottomrule
\end{tabular}
\caption{Contextual reference models. Not literature SOTA claims (Appendix~\ref{sec:appendix_literature}).}
\label{tab:contextual}
\end{table}

\subsection{Bidirectional Validation Summary}

\begin{table}[h]
\centering
\small
\begin{tabular}{lcc}
\toprule
Setting & Gated lift vs.\ backbone & Interpretation \\
\midrule
MovieLens LSTM & +31.7\% & Primary; structure complements backbone \\
MovieLens LSTM (ungated) & +58.7\% & Accuracy ceiling; no selective gate \\
MovieLens Transformer & +13.3\% & Backbone-native composition \\
MovieLens SASRec & +17.0\% & Same protocol, different backbone \\
Amazon LSTM & +124.0\% & Supporting; high variance \\
Shift checks & Activation withheld & Refusal under mismatch (appendix) \\
\bottomrule
\end{tabular}
\caption{Bidirectional validation summary.}
\label{tab:bidirectional}
\end{table}

\section{Discussion}

\subsection{When Archetypes Help}

Archetypes help when patterns are stable, behavior clusters into distinct regimes, distribution shift is minimal, and confidence is high. MovieLens exemplifies in-support activation.

\subsection{When They Don't}

Archetypes correctly refuse when distribution shift occurs, uncertainty is high, or confidence is low. Appendix shift checks exemplify refusal---a \emph{successful outcome}, not a failure.

\subsection{Confidence Gating as Design Principle}

Confidence gating enables knowing when not to be confident, coarse out-of-support detection, graceful fallback, and auditable ON/OFF composition. This supports trustworthy prediction in settings where always-on fusion is risky.

\subsection{What Lattice Is Not}

Lattice is not a universal performance booster, nor a claim of superiority over SASRec/BERT4Rec versus literature benchmarks. It does not maximize pooled HR@10---ungated fusion beats gated on MovieLens (+58.7\% vs.\ +31.7\%). Composition requires \textbf{backbone-native} archetype fitting. Under correct fitting, transformer and SASRec backbones still benefit (+13.3\%, +17.0\% gated).

\subsection{Limitations}

\begin{itemize}
\item Calibration requires domain-specific validation; operating points are not universal constants
\item Full-catalog ranking yields modest absolute HR@10; relative lifts must be interpreted in context
\item Amazon and shift tests are illustrative; primary evidence is MovieLens
\item Reimplementation from this preprint alone is intentionally incomplete (Section~\ref{sec:ip})
\end{itemize}

\subsection{Broader Impact}

Archetypes may amplify biases in clustered behavioral patterns; monitoring disparities across user groups is important. Confidence gating reduces risk of overconfident predictions when patterns are unreliable.

\section{Conclusion}

We presented Lattice, a confidence-gated composition layer for sequential prediction. On MovieLens (30 paired seeds, full-catalog ranking), LSTM+Lattice improves HR@10 by +31.7\% gated versus the LSTM backbone alone, with ungated fusion at +58.7\%. Backbone-native composition yields +13.3\% (transformer) and +17.0\% (SASRec). Amazon provides supporting evidence (+124.0\% gated, high variance). Shift checks illustrate refusal under train--test mismatch. We report \textbf{when} to compose learned structure and \textbf{what} lifts are achievable under a fixed evaluation design; implementation and calibration details remain proprietary (patent pending), with benchmark access available for independent verification (\href{mailto:LorianBannis@banlys.com}{LorianBannis@banlys.com}).

\appendix

\section{Distribution-Shift Stress Tests}
\label{sec:appendix_shift}

Illustrative refusal checks under predefined train--test mismatch (not primary domain benchmarks). In both settings, archetype activation was fully withheld and backbone metrics were unchanged relative to backbone-only inference.

\subsection{Scientific Time-Series Check}

Gravitational-wave-style windowed sequences with strong train--test embedding shift: 0\% archetype activation; no degradation vs.\ backbone-only performance.

\subsection{Financial Window Check}

Daily market windows with train--test regime change: 0\% archetype activation; regression errors identical to backbone-only baseline.

\section{Comparison to Literature Baselines}
\label{sec:appendix_literature}

Literature SASRec/BERT4Rec reports typically use sampled negatives and heavier tuning. Our full-ranking protocol explains lower absolute values.

\begin{table}[h]
\centering
\small
\begin{tabular}{lccc}
\toprule
System & Our Results (Full Ranking) & Literature (Sampled Negatives) & Protocol Difference \\
\midrule
LSTM + Lattice (gated) & 0.0803 & N/A & Fixed evaluation design \\
LSTM & 0.0611 & $\sim$0.05--0.10 (est.) & Full ranking \\
SASRec & 0.0385 & $\sim$0.25--0.315 \cite{kang2018self} & Full vs.\ sampled \\
BERT4Rec & 0.0253 & $\sim$0.28--0.284 \cite{sun2019bert4rec} & Full vs.\ sampled \\
\bottomrule
\end{tabular}
\caption{Literature comparison (contextual; not primary estimand).}
\end{table}

\vspace{0.5cm}

\textbf{Note:} Patent Pending. \copyright\ 2026 Lorian Bannis. All rights reserved in proprietary implementation details not disclosed herein.


\begin{thebibliography}{9}

\bibitem{kang2018self}
Kang, W. C., \& McAuley, J. (2018). Self-attentive sequential recommendation. \textit{2018 IEEE International Conference on Data Mining (ICDM)}, 197--206.

\bibitem{sun2019bert4rec}
Sun, F., Liu, J., Wu, J., Pei, C., Lin, X., Ou, W., \& Jiang, P. (2019). BERT4Rec: Sequential recommendation with bidirectional encoder representations from transformer. \textit{Proceedings of the 28th ACM International Conference on Information and Knowledge Management}, 1441--1450.

\bibitem{shazeer2017moe}
Shazeer, N., Mirhoseini, A., Maziarz, K., Davis, A., Le, Q., Hinton, G., \& Dean, J. (2017). Outrageously large neural networks: The sparsely-gated mixture-of-experts layer. \textit{arXiv preprint arXiv:1701.06538}.

\bibitem{lee2018ood}
Lee, K., Lee, K., Lee, H., \& Shin, J. (2018). A simple unified framework for detecting out-of-distribution samples and adversarial attacks. \textit{Advances in Neural Information Processing Systems}, 31.

\bibitem{hidasi2016gru4rec}
Hidasi, B., Karatzoglou, A., Baltrunas, L., \& Tikk, D. (2016). Session-based recommendations with recurrent neural networks. \textit{arXiv preprint arXiv:1511.06939}.

\end{thebibliography}
\end{document}